\documentclass[aoas2]{imsart}
\usepackage{amssymb}
\usepackage{amsmath}
\usepackage{algorithm}
\usepackage{algorithmic}
\usepackage{bbm}
\usepackage{color}
\usepackage{graphicx}
\usepackage{subfigure} 
\usepackage{natbib}

\usepackage[hidelinks]{hyperref}
\usepackage{url}

\newcommand{\ihat}{\mathbf {\hat \imath}}

\DeclareMathOperator*{\argmax}{arg\,max}

\renewcommand\Pr{\text{P}}
\newcommand{\ind}{\mathbbm{1}}

\begin{document}

\begin{frontmatter}

\title{Thompson sampling with the online bootstrap}
\runtitle{Thompson sampling with the online bootstrap}

\begin{aug}
\author{\fnms{Dean} \snm{Eckles}}
and
\author{\fnms{Maurits} \snm{Kaptein}\thanksref{t1}}
\affiliation{Facebook, Inc., and Radboud University, Nijmegen}
\runauthor{Eckles \& Kaptein}

\end{aug}

\begin{abstract}
Thompson sampling provides a solution to bandit problems in which new observations are allocated to arms with the posterior probability that an arm is optimal. While sometimes easy to implement and asymptotically optimal, Thompson sampling can be computationally demanding in large scale bandit problems, and its performance is dependent on the model fit to the observed data. We introduce bootstrap Thompson sampling (BTS), a heuristic method for solving bandit problems which modifies Thompson sampling by replacing the posterior distribution used in Thompson sampling by a bootstrap distribution. We first explain BTS and show that the performance of BTS is competitive to Thompson sampling in the well-studied Bernoulli bandit case. Subsequently, we detail why BTS using the online bootstrap is more scalable than regular Thompson sampling, and we show through simulation that BTS is more robust to a misspecified error distribution. BTS is an appealing modification of Thompson sampling, especially when samples from the posterior are otherwise not available or are costly.
\end{abstract}

\thankstext{t1}{Authors are listed alphabetically.}
\end{frontmatter}

\section{Introduction}
\label{intro}

Bandit problems, in which a specific action has an stochastic pay-off and the experimenter aims at maximizing the payoff over a sequence of recurring actions \citep{Whittle1980, Macready1998}, are prevalent. For example, in online advertising the action of selecting a specific ad out of a set of multiple ads for the current visitor of the website can be regarded a bandit problem: each ad has an uncertain payoff, and \emph{a priori} the ad with the highest pay-off is unknown \citep{Com2010}. The experimenter has to address exploration versus exploitation: presenting ads --- and observing the subsequent response --- about which little is known increases one's knowledge about the success rate of that ad. However, serving ads which one believes to be effective likely increases the overall payoff. Exploration and exploitation need to be balanced over the course of multiple interactions \citep{Audibert2009, Scott2010, Garivier2011}. Formally, bandit problems can be described as follows: at each time $t=1, \dots, T$, we have a set of possible actions $\mathcal{A}$. After choosing $a_t \in \mathcal{A}$ we observe reward $r_t$. The aim is to find a policy to select actions $a$ such that the cumulative reward $\mathcal{R}_c = \sum_{t=1}^{T} r_t$ is as large as possible. 

Many possible solutions to bandit problems have been suggested \citep[see, e.g.][]{Gittins1979, Whittle1980, Auer2010,Garivier2011}. Recently there has been substantial interest in \emph{Thompson sampling} \citep{THOMPSON1933} (or \emph{randomized probability matching} \citep{Scott2010}). In a theoretical analysis of Thompson sampling, \citet{kaufmann2012thompson} show that Thompson sampling for Bernoulli bandits asymptotically achieves the \citet{lai1985asymptotically} optimal performance limit. Empirical analysis of Thompson sampling, also in problems more complex than the Bernoulli bandit, shows a performance that is competitive to other solutions \citep{Chapelle2011, Scott2010}.

The basic idea of Thompson sampling is simple and intuitive: one randomly selects an action $a$ at time $t$ according to its estimated probability of being optimal (e.g., leading to the highest reward). Thompson sampling is formalized easily within a Bayesian framework \citep[cf.][]{Scott2010}. The set of past observations $\mathcal{D}$ consists of the actions $a_{(1, \dots, t)}$ and the rewards $r_{(1, \dots, t)}$. The rewards are modeled using a parametric likelihood function: $\Pr(r | a, \theta)$ where $\theta$ is a set of parameters. Using Bayes rule it is, in some problems, easy to compute or sample from $\Pr(\theta | \mathcal{D})$. Given that we can compute $\Pr(\theta | \mathcal{D})$ we can select an action according to its probability of being optimal:
\begin{equation}
\label{eq:Thomp:1}
\begin{aligned}
\int \ind \left[\mathbb{E}(r|a, \theta) = \underset{a'}{\max} \: \mathbb{E}(r | a', \theta)\right] \Pr(\theta | \mathcal{D}) d\theta
\end{aligned}
\end{equation}
where $\ind$ is the indicator function. In practice it is not necessary to compute the above integral: it suffices to draw a random sample $\theta^*$ from the posterior at each round and select the action with the highest estimated reward given the current draw. 

When it is easy to sample from $\Pr(\theta | \mathcal{D})$, Thompson sampling is easy to implement. However, to be practically feasible for many problems, and thus scalable to large $T$ or to complex likelihood functions, Thompson sampling requires computationally efficient sampling from $\Pr(\theta | \mathcal{D} )$. In practice $\Pr(\theta | \mathcal{D} )$ might not always be easily available: already in situations in which a logit or probit model is used to model the expected reward of the actions, $\Pr(\theta | \mathcal{D} )$ is not available in closed form and is then often computed using MCMC methods, which can be computationally costly. Furthermore, for many likelihood functions it is hard to update the posterior online (i.e., row-by-row) thus requiring inspection of the full dataset $\mathcal{D}$ at each iteration. Both of these properties make that for a number of applied problems the scalability of Thompson sampling might be limited. Also, Thompson sampling is a parametric method, so its performance depends on the accuracy of the model that is used to compute $\Pr(r | a, \theta)$.
Thus, Thompson sampling may not be very robust to common forms of model misspecification.

To address the problems of scalability and robustness of Thompson sampling encountered in complex bandit problems,
we introduce a modification of Thompson sampling that we call \emph{bootstrap Thompson sampling} (BTS).
BTS replaces the posterior $\Pr(\theta | \mathcal{D})$ by a bootstrap distribution of the point estimate $\hat{\theta}$.
Some bootstrap methods are especially computationally appealing. In particular, bootstrap methods that involve randomly reweighting data \citep{rubin_bayesian_1981}, rather than resampling data, can be conducted online \citep{lee_lossless_2004,Owen2012,oza_bagging_2001}. For BTS we use a bootstrap method in which, for each bootstrap replicate $j \in \{1, \dots, J\}$, each observation gets a weight $w_{tj} \sim 2 \times \text{Bernoulli}(1/2)$. Following \citet[][\S 3.3]{Owen2012}, we refer to this bootstrap as the \emph{double-or-nothing bootstrap} (DoNB) or \emph{online half-sampling}.\footnote{Since the absolute scale of the weights does not matter for most estimators, it is equivalent to have the weights be 0 or 1, rather than 0 or 2. Other weight distributions could be used for various reasons. For example, using exponential weights is the so-called Bayesian bootstrap \citep{rubin_bayesian_1981}. In that case, each observation has positive weight in each replicate, which can avoid numerical problems in some settings, but requires updating all replicates for each observation. \citet[][\S 3.3]{Owen2012} compare weight distributions for the bootstrapping the sample mean.}

Statisticians have noted relationships between bootstrap distributions \citep{efron1979bootstrap} and Bayesian posteriors.
With a particular weight distribution and nonparametric model of the distribution of observations, the bootstrap distribution and the posterior coincide \citep{rubin_bayesian_1981}.
In other cases, the bootstrap distribution $\tilde{\theta}$ can be used to approximate a posterior \citep[e.g.,][]{Efron2011, newton_approximate_1994}, e.g., as a proposal distribution in importance sampling.
Moreover, sometimes the difference between the bootstrap distribution and the Bayesian posterior is that the bootstrap distribution is more robust to model misspecification, such that if they differ substantially the bootstrap distribution may even be preferred \citep{liu_approximate_1994, szpiro2010}. 

In the remainder of this article we first illustrate BTS using a simple $K$-armed Bernoulli bandit and show its competitive performance compared to Thompson sampling. In this section we also discuss in more detail the choice of $J$ which can be regarded a tuning parameter in BTS. Subsequently, we discuss the scalability of BTS by analyzing its computational complexity, and we examine empirically the robustness of BTS in cases of model misspecification. 

\section{Illustrating BTS using the $K$-armed Bernoulli bandit problem}
\label{sec:k-arm-thompson}

A commonly used example of a bandit problem is the $K$-arm Bernoulli bandit problem, where $r_t \in \{0,1\}$, and the action $a$ is to select an arm $i=1, \dots, K$ at time $t$. The reward of the $i$-th arm follows a Bernoulli distribution with true mean $\theta_i$. The implementation of standard Thompson sampling using Beta priors for each arm is straightforward: For each arm $i$ one sets up a Beta-Bernoulli model and at each round one obtains a single draw $\theta^*_i$ from each of the Beta posteriors, plays the arm $\ihat = \argmax_i \theta^*_i$, and subsequently uses the observed reward $r_t$ to update the Beta posterior; for a full description see \cite{Chapelle2011}. The BTS implementation is similar, but instead of using a Beta-Bernoulli model to compute $\Pr(\theta | \mathcal{D})$, we use the DoNB bootstrap to obtain a sample from the bootstrap distribution $\tilde{\theta}$; that is, from each $\tilde{\theta}_i$ we obtain a draw $\theta^*_i$ and again play the arm $\ihat = \argmax_i \theta^*_i$.

\begin{figure*}[t!]
 \centering
 \includegraphics[width=.95\textwidth]{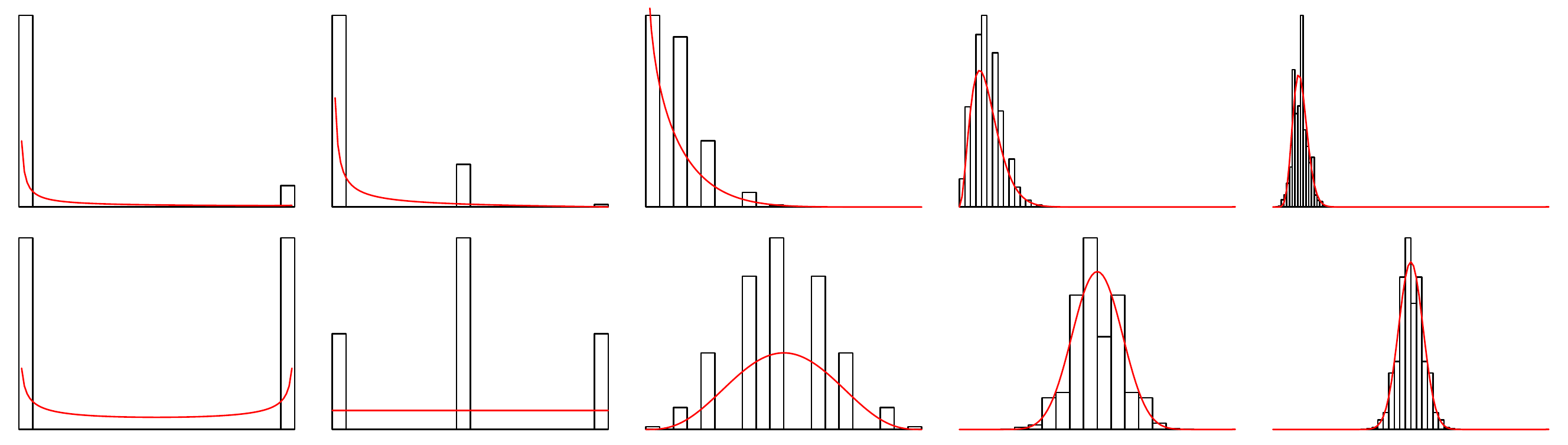}
  \caption{Illustration of the ``evolution'' of the marginal bootstrap distribution $\tilde{\theta}$ for increasing $n$ in the Bernoulli case with true parameter $\theta = .1$ (top row) and $\theta = .5$ (bottom row). Presented are, from left to right, the theoretical distribution for $n \in \{1,2,8,32,128\}$. Superimposed in red is the expected $\text{Beta}(\alpha=\theta n, \beta=(1-\theta)n)$ used for standard Thompson sampling.}
 \label{fig:boot-post}
\end{figure*}

To illustrate, Figure \ref{fig:boot-post} presents the theoretical expected bootstrap distribution $\tilde{\theta}$ for a true $\theta \in \{.1, .5\}$ for increasing numbers of observations $n$. For the first observation ($n=1$) $\tilde{\theta}$ takes on values $0$ (with probability $1-\theta$) and $1$ (with probability $\theta$). For $n=2$, possible values are $0$, $\frac{1}{2}$, and $1$ with probabilities proportional to $\theta^2$, $\theta(1-\theta)$, and $(1-\theta)^2$. As the sequence continues the number of possible values increases, and the probability mass centers around the true $\theta$. 

Our implementation of BTS for the $K$-armed Bernoulli bandit is given in Algorithm \ref{Alg:BTS_Bernoulli}. We start with an initial belief about the parameter by setting $\alpha_{ij} = 1$ and $\beta_{ij} = 1$ for each arm $i$ and each bootstrap replicate $j$.
To decide on an action, for each arm $i$, we uniformly randomly draw one of the $J$ replicates $j_i$, and play arm $\ihat$ with the largest point estimate $\hat{\theta}_{i j_i} = \alpha_{i j_i} / (\alpha_{i j_i} + \beta_{i j_i})$, breaking ties randomly. After observing reward $r_t$, we update each bootstrap replicate with probability 0.5. 

\begin{algorithm}
\caption{The BTS solution for the Bernoulli bandit}
\label{Alg:BTS_Bernoulli}
\begin{algorithmic}
\REQUIRE $\alpha$, $\beta$ prior parameters for bootstrap Thompson sampling.
\STATE $\alpha_{ij} := \alpha$, $\beta_{ij} := \beta$ \{For each arm $i$ and each bootstrap replicate $j$\}
\FOR{$t=1, \dots, T$}
	\FOR{$i, \dots, K$}
		\STATE Sample $j_i$ from uniform $1, \dots, J$ bootstrap replicates
		\STATE Retrieve $\alpha_{ij_i}, \beta_{ij_i}$		
	\ENDFOR
	\STATE Play arm $i = \argmax_i{\: \alpha_{i j_i} / (\alpha_{i j_i} + \beta_{i j_i})}$ and observe reward $r_t$
	\FOR{$j, \dots, J$}
		\STATE Sample $d_{j}$ from $\text{Bernoulli}(1/2)$
		\IF{$d_j=1$}
			\STATE $\alpha_{\ihat j} = \alpha_{\ihat j} + r_t$
			\STATE $\beta_{\ihat j} = \beta_{\ihat j} + (1 - r_t)$
		\ENDIF
	\ENDFOR
\ENDFOR
\end{algorithmic}
\end{algorithm}

The choice of $J$ limits the number of samples we have from the bootstrap distribution $\tilde{\theta}$. For small $J$, BTS is expected to become greedy: if in all combinations of the $J$ replicates, some arm $i$ does not have the largest point estimate, then this arm has zero probability of being played until this changes. In such a case, BTS could tend to over-exploit. A large $J$, while computationally more costly, will allow for more exploration.

\begin{figure*}[t!]
 \centering
 \includegraphics[width=1\textwidth]{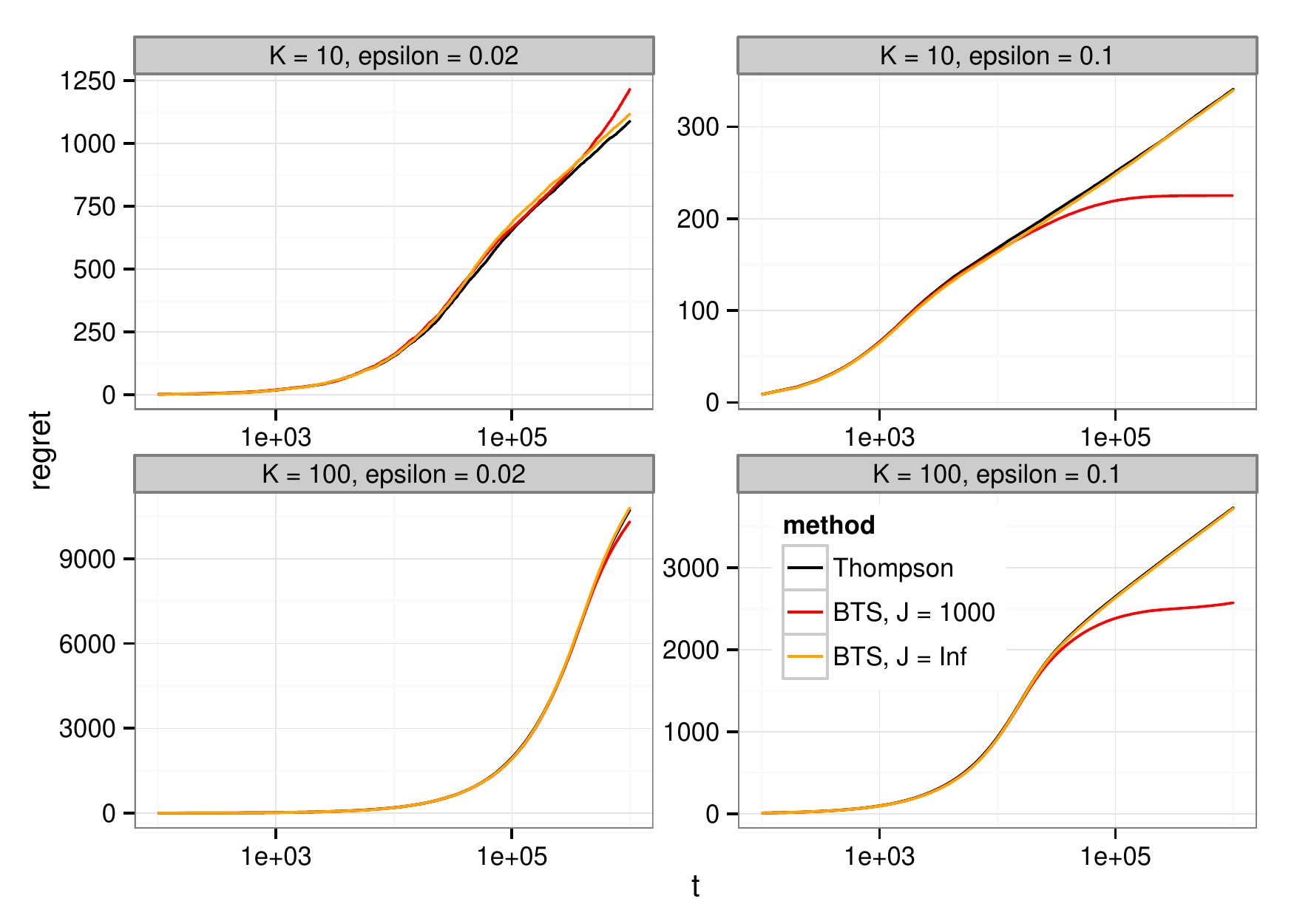}
  \caption{
Empirical regret for Thompson sampling and BTS in a $K$-armed binomial bandit problem with varied differences between the optimal arm and all others $\epsilon$.
For BTS with $J = 1000$ bootstrap replicates, Algorithm \ref{Alg:BTS_Bernoulli} is used.
BTS with $J = 1000$ sometimes shows lower mean regret than Thompson sampling when the arms are more different (i.e., $\epsilon = 0.1$).
This lower empirical regret is because the finite number of bootstrap replicates $J$
result in a method that is greedier than Thompson sampling.
For comparison, we show the performance when $J$ is effectively infinite (see main text). In this case, BTS and Thompson sampling have very similar performance.
}
 \label{Fig:Regret-Binomial}
\end{figure*}

To examine the performance of BTS we present an empirical comparison of Thompson sampling to BTS in the $K$-armed Bernoulli bandit case. In our simulations the best arm has a reward probability of $.5$, and the $K-1$ other arms have a probability of $.5-\epsilon$.
We examine cases with $K \in \{10, 100\}$
and $\epsilon \in \{.02, .1\}$.
Figure \ref{Fig:Regret-Binomial} presents the empirical regret over time, $R_{t} = .5 t - \sum_{t=1}^{t}(r_{t'})$, of Thompson and BTS.\footnote{To decrease simulation error, in this computation we replace $.5 t$ with the observed reward for playing the optimal arm with the same random numbers.}
The results presented here are for $t = 1, \dots, T=10^6$.
Figure \ref{Fig:Regret-Binomial} presents the average regret over $1000$ simulation runs with $J=1000$ bootstrap replicates. The mean regret of BTS is similar to that of standard Thompson sampling. In some sets of simulations, BTS has lower empirical regret than Thompson sampling; however, this is because the use of a finite $J$ makes BTS greedy. For comparison, we present a version of the BTS algorithm in which a new bootstrap replicate is constructed for each $t$ such that $J$ is effectively infinite.\footnote{We implement this by storing the sufficient statistics (the number of successes and failures) and resampling these at each round according to the DoNB bootstrap. This is possible in simple bandit problems such as the $K$-armed Bernoulli bandit case or when the number of unique combinations of arms and rewards is small.}
This version of BTS has performance very closely comparable to Thompson sampling.

\begin{figure*}[t!]
 \centering
 \includegraphics[width=.6\textwidth]{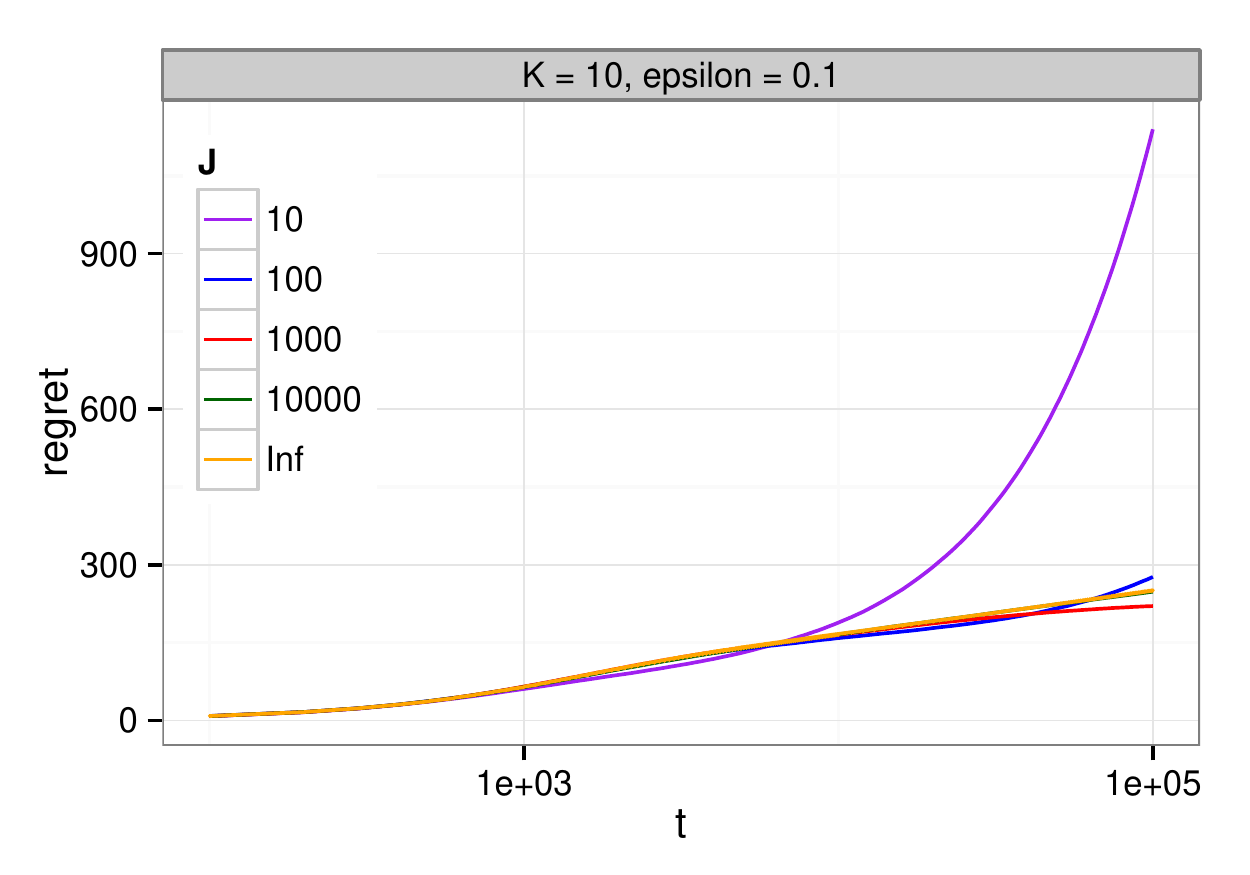}
  \caption{
Comparison of empirical regret for BTS with varied number of bootstrap replicates $J$.
Using a very small $J$ (e.g., 10) results in an over-greedy method that can get ``stuck'' on non-optimal arms, since the optimal arm may not win in any of the $J$ replicates. Larger values of $J$ give performance much more similar to that when $J$ is effectively infinite.
}
 \label{Fig:Regret-Binomial-J}
\end{figure*}

To further examine the importance of the number of bootstrap replicates $J$, Figure \ref{Fig:Regret-Binomial-J} presents the cumulative regret for the $K=10$ and $\epsilon = .1$ with $J \in \{10,100,1000,10000, \infty \}$.
Here it is clear that in cases when $J$ is small, the algorithm becomes too greedy and thus, in the worst case, suffers linear regret. $J$ can be thought of as a tuning parameter for the BTS algorithm: with large $\epsilon$ one might settle for lower values of $J$ since arms are more easily separable and the chance of getting ``stuck'' on an incorrect arm is small (albeit still positive). If small values of $\epsilon$ are of interest then a higher number of bootstrap replicates will be necessary. Similarly, if in a practical application the horizon $T$ is comparatively small, a small number of bootstrap replicates suffices: the performance of BTS before becoming greedy is similar to Thompson sampling. The tuning parameter $J$ can thus also be evaluated in relation to the expected $T$ in applied problems where for large $T$ more replicates are necessary. The properties of the theoretical bootstrap distribution, $\tilde{\theta}$, for this purpose may need further analytical scrutiny.

\section{Scalability of BTS compared to Thompson sampling}

Figure \ref{Fig:Regret-Binomial} showed the competitive performance of BTS to Thompson sampling in the Bernoulli bandit case and highlighted the influence of $J$ as a tuning parameter for the desired precision given the problem at hand. However, since for the Bernoulli Bandit computation of $\Pr(\theta | \mathcal{D})$ is straightforward, and easily done online, this illustrative example does not motivate using BTS for reasons of scalability.

For larger problems computation of $\Pr(\theta | \mathcal{D})$ will not aways be straightforward, which is partly our motivation to present BTS.
Consider, for example, the generalization of the simple bandit problem to a contextual bandit problem: here the set of past observations $\mathcal{D}$ is composed of a triplet $\{z, a, r\}$, where the $z$ denotes the context: additional information not represented within the action itself that is observed at a specific time, rather than assigned by the experimenter. Equation \ref{eq:Thomp:1} now becomes
\begin{equation}
\label{eq:Thomp:1}
\begin{aligned}
\int \ind \left[\mathbb{E}(r | a, z, \theta) = \underset{a'}{\max} \: \mathbb{E}(r | a', z, \theta)\right] \Pr(\theta | \mathcal{D}) d\theta.
\end{aligned}
\end{equation}
In this specification, and with rewards $r_t \in \{0,1\}$, one would often setup a logit or probit model to sample from $\Pr(\theta | \mathcal{D})$. No general closed form for $\Pr(\theta | \mathcal{D})$ exists. Thus one would resort to MCMC methods. At each time $t$ when a decision need to be made this can be computationally costly since the chain has to converge, but more cumbersome is the fact that no online update is available. To produce a sample from the posterior at time $t$ the algorithm will revisit all data $\mathcal{D}_{(t=1, \dots, t=t)}$ giving a computational complexity of $O(t)$ at each update.\footnote{This specification removes all the constants required, e.g., for MCMC chains to converge. In some cases, this might also depend on $t$, potentially increasing complexity. Note that faster algorithms might be available in special cases and through the use of other methods for approximating the likelihood or posterior, to which BTS might be viewed as another competitor. In fact, further elaborations on BTS, such as the use of the bootstrap distribution as a proposal distribution \citep{newton_approximate_1994}, may sometimes be alternatives. However, many presentations of Thompson sampling make use of conjugacy relationships, MCMC, or problem-specific formulations \citep[e.g.][]{Chapelle2011, Scott2010}.}

For BTS however, as long as $\hat{\theta}$ can be computed online, which is often possible even when $\Pr(\theta | \mathcal{D})$ cannot be updated and sampled from online, the computational complexity of getting an updated sample at time $t$ is $O(J) = O(1)$ since $J$ is a constant. Thus, for bandit problems, especially such as cases as contextual bandit problems, Thompson sampling can be cumbersome computationally while BTS can be computed fully online. BTS is thus sometimes a more scalable alternative to Thompson sampling.

\section{Robustness of BTS compared to Thompson sampling}

Besides scalability of BTS compared to Thompson sampling, we expected BTS to be more robust to some kinds of model misspecification, given the robustness of the bootstrap for statistical inference \citep[cf.][]{liu_approximate_1994}. To empirically examine this we setup a simulation in which we compare the performance of BTS with Thompson sampling in cases of heteroscedastic Gaussian errors. Bootstrap methods are often used in statistical inference for regression coefficients and predictions when errors may be heteroscedastic, including because the model fit for the conditional mean may be incorrect \citep{freedman1981bootstrapping}.
The relatively simple data-generating process has three factors, $x_t = \{x_1, x_2, x_3 \}$, with two levels $l \in \{0,1\}$ each. Thus, in our simulation $a \in \{1, \dots, 8 \}$ referring to all $2^3$ possible configurations. The true data generating model is $y = \mathbf{X} \beta + \epsilon$ where $\epsilon \sim \mathcal{N}(0, \mathbf{X} \sigma^2)$. We use

\begin{equation}
\mathbf{X} = 
\begin{pmatrix}
1&0&0&0&0&0&0&0 \\
1&1&0&0&0&0&0&0 \\
1&0&1&0&0&0&0&0 \\
1&1&1&0&1&0&0&0 \\
1&0&0&1&0&0&0&0 \\
1&1&0&1&0&1&0&0 \\
1&0&1&1&0&0&1&0 \\
1&1&1&1&1&1&1&0 \\
1&1&1&1&1&1&1&1 \\
\end{pmatrix}, \; \;
\beta = 
\begin{pmatrix}
1.00 \\
-0.20 \\
0.10 \\
0.20 \\
0.10 \\
0.05 \\
0.10 \\
0.01 \\
\end{pmatrix}, \; \;
\sigma^2 = 
\begin{pmatrix}
1 \\
0 \\
0\\
\gamma \\
0 \\
0 \\
0 \\
\gamma 
\end{pmatrix}
\end{equation}
where $\mathbf{X}$ is the design matrix, with each row corresponding to one of the 8 arms, $\beta$ denotes the vector of coefficients for the linear model including all interactions. Finally, we use $\sigma^2$ to denote the vector of variance components for each column of $\mathbf{X}$. We vary $\gamma$ to create different degrees of heteroscedasticity. Note that arm 7, with an expected reward 1.40, is the optimal arm. The next best arm is arm 8 with expected reward 1.36, while arm 2, with an expected reward 0.8, is the worst arm. 

We then compare Thompson sampling and BTS. For Thompson sampling, we fit a Bayesian linear model using a Gaussian prior with variance 1. We fit the linear model each time to the full dataset $\mathcal{D}$ consisting of $r_1, \dots, r_t', x_1, \dots, x_t'$ where $x_t$ denotes the feature vector at time $t$. Next, we take a random draw from $\Pr(\theta | \mathcal{D})$ to facilitate standard Thompson sampling. For BTS, we use a fully online version using the well known online (summation form) implementation of a linear OLS model \cite[See][p. 3 for a worked out version]{Chu2007} with a ridge penalty $\lambda = 1$.\footnote{When the error variance is 1, the ridge point estimate is equivalent to the posterior mode with a Gaussian prior with variance 1 \citep[][\S 3.4.1]{hastie_elements_2008}.
}
Here we compute in online for each selected DoNB replicate $j \in \{1 , \dots , J\}$ matrix $\mathbf{A} = \sum_{t=1}^T x_t x_t^T$ and vector $b = \sum_{t=1}^T x_t y_t$. We then select a random $j$, compute $\theta^* = A^{-1} b$ and play the arm that maximizes the reward given this estimate of the model coefficients. In the simulation study we use $J=1000$ to approximate the bootstrap distribution $\tilde{\theta}$.

\begin{figure*}[t!]
 \centering
 \includegraphics[width=0.7\textwidth]{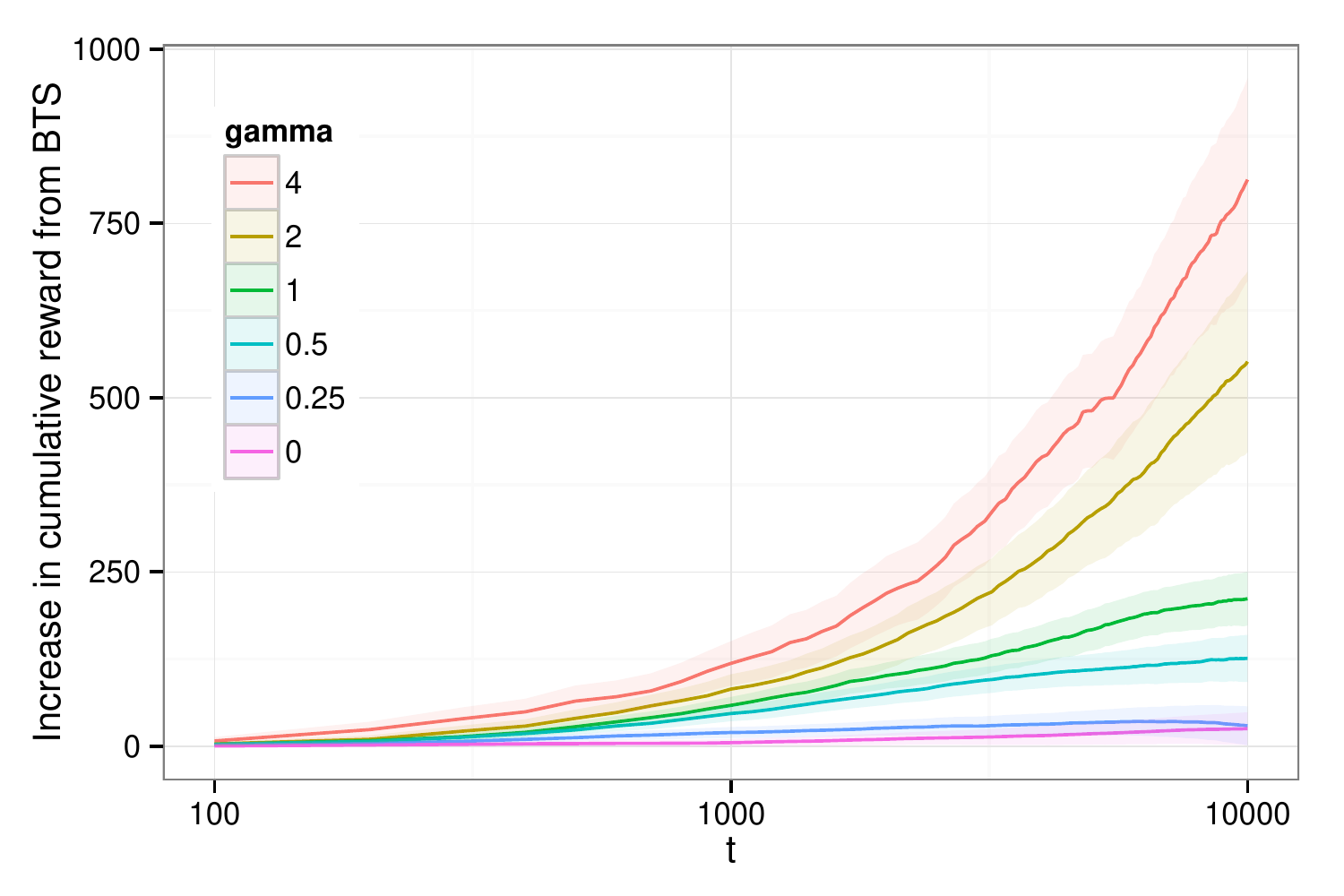}
  \caption{Comparison of Thompson sampling and BTS (with $J = 1000$) with a factorial design and continuous rewards with a heteroscedastic error distribution. Lines correspond to differing degrees of heteroscedasticity for $\gamma \in \{0, .25, .5, 1\}$. Increasing heteroscedasticity produces a larger differences between Thompson sampling and BTS. Bands are point-wise 95\% confidence intervals using a normal approximation.}  
 \label{Fig:Regret-Factorial}
\end{figure*}

We examine a range of cases in which the model has a misspecified error distribution. We fit a full factorial model, but we ignore the heteroscedasticity present in the data-generating model. It is well known that ignoring such heteroscedasticity will often be anticonservative; in this Bayesian context, the posterior for some arms would often be more concentrated than with a more flexible model.\footnote{There is recent work in developing Bayesian methods that allow for heteroscedasticity, including that resulting from model misspecification \citep[e.g.,][]{szpiro2010}.}
Figure \ref{Fig:Regret-Factorial} presents the difference in cumulative reward between BTS and Thompson sampling for $t = 1, \dots, 10^4$ for varying degrees of heteroscedasticity, $\gamma \in \{0, .25, .5, 1, 2, 4\}$, with 100 simulations. Even with a relatively small degree of misspecification (e.g., $\gamma = 0.5$) and with small $t$ (e.g., $t = 1000$), BTS has significantly higher cumulative reward (lower cumulative regret) than Thompson sampling. As expected, this difference increases with $\gamma$.

\section{Discussion}

In this paper we introduced BTS as an alternative and computationally efficient substitute for Thompson sampling. BTS relies on the same idea as Thompson sampling: to optimize the exploration--exploitation trade-off a competitive method is to play action $a$ with the probability of it being the best action. However, where Thompson sampling relies on a fully Bayesian specification to sample from $\Pr(\theta | \mathcal{D})$ we substitute this latter distribution by the bootstrap distribution $\tilde{\theta}$.
By a reweighting bootstrap, such as the the double-or-nothing bootstrap used here, BTS can be implemented fully online whenever the point estimate can be implemented online.

The theoretical appeal of BTS can be motivated from relationship of the bootstrap distribution to the Bayesian posterior or the true sampling distribution for the parameter \citep[cf.][]{Efron2011}. We have only referred to and illustrated such a comparison (e.g., in Figure \ref{fig:boot-post}). This relationship needs further scrutiny from the perspective of how differences between these distributions matter for the asymptotic behavior of BTS as a policy for bandit problems. However, the current empirical evaluation warrants additional attention to BTS as a solution to bandit problems. Bootstrap methods are also often used in statistical inference when observations are dependent (e.g., time series, random effects models); this suggests the value of future theoretical and empirical analysis of BTS with dependent data (e.g., multiple observations of the same person) when using an appropriate bootstrap method (e.g., cluster bootstrap).

In practical appeal of BTS is in part motivated by its computational advantages. The computational demands of the many MCMC approaches to sample from $\Pr(\theta | \mathcal{D})$ as needed for Thompson sampling quickly increases as $\mathcal{D}$ grows in size (e.g., $t$ becomes large). The computation required for each round of BTS, however, need not depend on $t$ and thus can be feasible even when $t$ gets extremely large. This makes online BTS a good candidate for many real explore--exploit problems where a point estimate of $\theta$ can be obtained online, but $\Pr(\theta | \mathcal{D})$ is hard to compute. Besides the clear differences computational complexity, BTS is also appealing for large-scale problems because the procedure is easily parallelized: it is straightforward to distribute the computation of bootstrap replicates $J$ over multiple cores or machines. For example, users of a Web service can be routed to different prediction providers, each of which has a different set of bootstrap replicates. When the rewards are later observed, each can be added to the data for each replicate with probability $1/2$.

We presented a number of empirical simulations of the performance of BTS in the canonical Bernoulli bandit and in a Gaussian bandit problem. The Bernoulli bandit simulations allowed us to demonstrate the competitive performance of BTS, and highlighted the importance of tuning parameter $J$. The Gaussian bandit illustrated robustness of BTS in cases of model misspecification.
We conclude that BTS is competitive in performance, more amenable to some large-scale problems, and, at least in some cases, more robust than Thompson sampling.
The observation that BTS can over-exploit when the number of online bootstrap replicates is too small needs further attention. The number of bootstrap replicates $J$ can be regarded a tuning parameter in applied problems --- just as the posterior in Thompson sampling can be artificially concentrated \citep{Chapelle2011} ---
or could be adapted dynamically as the estimates of the arms evolve.
We should develop a better understanding of the relationship between the number of bootstrap replicates and the tendency to favor exploitation over exploration. Finally, future work should develop more deeply the analytical properties of BTS and make further comparison to other methods for approximating the likelihood or posterior in scalable and parallelizable ways.

\subsection*{Acknowledgements}
This work was improved by comments from Eytan Bakshy, Thomas Barrios, Daniel Merl, members of the Facebook Core Data Science team, and anonymous reviewers.

\subsubsection*{References}
\bibliography{library}
\bibliographystyle{apa}
\end{document}